\documentclass[conference]{IEEEtran}
\IEEEoverridecommandlockouts
\usepackage{cite}
\usepackage{amsmath,amssymb,amsfonts}
\usepackage{algorithmic}
\usepackage{graphicx}
\usepackage{textcomp}
\usepackage{xcolor}
\usepackage{graphics,cite,graphicx,amsfonts,flafter}
\usepackage{amsmath,amssymb,url,verbatim,graphicx}
\usepackage{subfig,float,cases,mathptmx,epsfig,epsf,wrapfig}
\usepackage{xfrac,caption,array,supertabular}
\usepackage{textcomp,multirow,multicol}
\usepackage{latexsym,siunitx,booktabs}
\usepackage[section]{placeins}
\usepackage[linesnumbered,ruled,vlined]{algorithm2e}
\usepackage{xcolor}
\usepackage{bbm}
\usepackage{breqn}
\renewcommand{\baselinestretch}{1}

\def\be{\begin{eqnarray}}
\def\ee{\end{eqnarray}}
\def\ben{\begin{eqnarray*}}
\def\een{\end{eqnarray*}}

\def\bc{\begin{center}}
\def\ec{\end{center}}

\newcommand{\bX}{{\textbf{X}} } 
\newcommand{\bx}{{\textbf{x}} } 
\newcommand{\by}{{\textbf{y}} } 
 
\usepackage{titlesec}    
\begin{document}

\title{Bayesian Nonparametric View to Spawning }

\author{\IEEEauthorblockN{Bahman Moraffah}
\IEEEauthorblockA{\textit{School of Electrical, Computer and Energy Engineering} \\
\textit{Arizona State University}\\
Tempe, AZ \\
bahman.moraffah@asu.edu}}

\maketitle

\begin{abstract}
In tracking multiple objects, it is often assumed that each observation (measurement) is originated from one and only one object. However, we may encounter a situation that each measurement may or may not be associated with multiple objects at each time step --spawning. Therefore, the association of each measurement to multiple objects is a crucial task to perform in order to track multiple objects with birth and death. In this paper, we introduce a novel Bayesian nonparametric approach that models a scenario where each observation may be drawn from an unknown number of objects for which it provides a tractable Markov chain Monte Carlo (MCMC) approach to sample from the posterior distribution. The number of objects at each time step, itself, is also assumed to be unknown. We, then, show through experiments the advantage of nonparametric modeling to scenarios with spawning events. Our experiment results also demonstrate the advantages of our framework over the existing methods. 
\end{abstract}


\section{Introduction}
Multiple object tracking with spawning events has become ubiquitous in recent years. However, due to the nature of observations that may or may not be generated from multiple time-varying objects, the association of the observations to the corresponding objects as well as tracking the time-varying number of objects are difficult and computationally demanding tasks. In this paper, we exploit the Bayesian nonparametric modeling over multiple objects as well as nonparametric modeling for observation association. Bayesian nonparametric frameworks have shown promising results in recent years. Due to the efficiency of Bayesian nonparametric modeling, in  \cite{ caron2012, moraffah2019random, moraffah2018,  moraffah2019PY}, authors employed nonparametric Bayesian modeling to improve the tracking of multiple objects. Hierarchical Bayesian nonparametric modeling has been exploited for switching in linear dynamic systems \cite{Fox2011} and for tracking an object with multiple observation sources \cite{moraffah2019-mdm, moraffah2020bayesian}.  In \cite{moraffah2019bayesian_thesis, moraffah2019inference-arxiv}, authors provide mathematical guarantees as well as Gibbs sampling formulations for the dependent nonparametric models. Moreover, Bayesian nonparametric frameworks are introduced for tracking objects in high clutter \cite{moraffah_clutter}. It is worth pointing out that even though all these frameworks provide auspicious results that are shown to outperform the previously introduced methods, they all assume each measurement is originated from one and only one object. This assumption introduces extensive challenges in applications that may violate this premise for which each measurement may or may not be originated from more than one object. For instance, the sudden creation of many objects due to fragmentation may potentially cause huge risks for tracking the objects. In this case, a simple tracking model cannot provide an accurate estimate of the object's trajectory.  In recent years, authors proposed solutions for spawning through the random finite set (RFS) theory \cite{bar1990tracking, Faber2016}. Zero-Inflated Poisson spawn model is introduced which is derived from cardinalized probability hypothesis density (CPHD) filter \cite{Bryant17}. This model only accounts for survival and birth. In \cite{Bryant18}, the generalized labeled multi-Bernoulli (GLMB)-based filter is used to address the birth and death issue posed by the CPHD-based technique. These models utilize Bayesian analysis on random sets and introduce Bayesian filtering techniques. However, It is shown that RFS-based solutions are computationally demanding. In this paper, we introduce Bayesian nonparametric techniques based on the Dependent Dirichlet process (DDP) for multiple object tracking and the Indian Buffet Process (IBP) to account for both tracking time-varying number of objects and spawning. The nature of the IBP prior allows each measurement to come from multiple objects. Our nonparametric spawning model accounts for the measurement association to multiple objects where each measurement may be originated from more than one object. Moreover, the number of objects that originate a measurement is assumed to be unknown.   In this paper, we propose a novel Bayesian nonparametric approach that not only enables the joint estimation of a spawned object's state as well as the object cardinality but also addresses the measurement to objects association.  We incorporate a dependent process to account for multiple object tracking introduced in \cite{moraffah2018} and the IBP-based technique to address spawning.

The rest of the paper is organized as follows. In section \ref{sec:background}, we provide a quick review of the Dependent Dirichlet process (DDP) and the Indian Buffet Process (IBP). Section \ref{sec:PF} and section \ref{sec:proposed_method} state the problem formulations and our proposed framework, respectively. In section \ref{sec:ex}, we then demonstrate through experiments that our proposed framework outperforms the GLMB-based spawning technique introduced in \cite{Bryant18}. 

\textbf{Notation:} We denote the the $j$th observation and the $\ell$th object state at time $k$ by $\by_{j,k}$ and $\bx_{\ell,k}$, respectively. The set of all possible states at time $k$ is shown by $\mathbb{X}_k$. Parameters associated with states and observations are $\theta$ and $\phi$, respectively. We indicate the $j$th element of vector $V$ by $[V]_j$. We assume exchangeability throughout this paper. 

\section{Background}
\label{sec:background}
\subsection{Dependent Dirichlet Process (DDP)}

In recent years, the use of Dirichlet Processes (DPs) for modeling complex data has been well-established. However, DPs may not capture the dependency in data. To address the dependency issue, MacEachern introduced dependent nonparametric Bayesian processes as  \cite{maceachern1999dependent} 
\be
G_x(\cdot) = \sum_{i=1}^\infty \pi_{i,x}\delta_{\theta{i,x}}
\ee
where $\pi_{i,x} = v_{i,x}\prod_{j=1}^{i-1}(1-v_{j,x})$. In this setup,  $v_{i,\bX}$ is considered to be a stochastic process. It is worth emphasizing that unlike vanilla Dirichlet process, in this formulation both $v_{i,\bX}$ and $\theta_{i,\bX}$ are stochastic processes. Moraffah in \cite{moraffah2018} introduced a class of Dependent Dirichlet process prior that can be employed in tracking multiple time-varying objects. This formulation captures birth, death, and remaining of time-varying objects in the field of view. Taking advantage of the Chinese Restaurant Process (CRP) representation of DP, the $\text{DDP}(\alpha)$ prior is constructed analogously as follows:
\begin{itemize}
\item \textbf{Case 1:}  Assigning the $\ell$th object at time $k$  to one of the survived and transitioned to time $k$ which have been occupied by at least one of the first $(\ell - 1)$ w.p.
\be
\begin{split}
\Pi_1 (\text{$j$th cluster}|\theta_{1,k},\dots,&\theta_{\ell-1,k}) = \\
&\frac{[V_{k|k-1}]_j + [V_k]_j}{\alpha+\sum_i[V_{k|k-1}]_i + [V_k]_i }
\end{split}
\ee
where $V_{k|k-1}$ and $V_k$ are the vector of the number of objects in each cluster after transitioning to time $k$ and at time $k$, respectively. 
\item \textbf{Case 2:} Assigning the $\ell$th object to one of the survived and transitioned cluster at time $k$ where it has not been assigned to any of the previous objects w.p.
\be
\begin{split}
\Pi_1 (\text{$j$th cluster}|\theta_{1,k},\dots,&\theta_{\ell-1,k}) = \\
&\frac{[V_{k|k-1}]_j }{\alpha+ \sum_i  [V_{k|k-1}]_i + [V_k]_i }
\end{split}
\ee
\item \textbf{Case 3:} Assigning a new cluster to the $\ell$ object w.p.
\be
\begin{split}
\Pi_1 (\text{new cluster}|\theta_{1,k},\dots,&\theta_{\ell-1,k}) = \\
&\frac{\alpha}{\alpha+ \sum_i[V_{k|k-1}]_i + [V_k]_i }.
\end{split}
\ee
\end{itemize}

\subsection{Indian Buffet Process (IBP)}
Indian Buffet Process (IBP) is a class of nonparametric models that can be considered as a prior on the sparse infinite binary matrices \cite{JMLR:v12:griffiths11a}. Unlike CRP which is the distribution on the partitions of data induced by DP-- implying each data point belongs to one and only one cluster-- the IBP provides distributions over infinite binary matrices where each data may comprise more than one feature. Using CRP analogy, the IBP prior is constructed as follows \cite{JMLR:v12:griffiths11a}:
\begin{itemize}
\item The first customer starts at the left of the buffet, and takes food from each dish, stopping after a Poisson($\gamma$) number of dishes.
\item The $j$th customer samples dishes proportional to their popularity w.p. $n_i/j$ and tries Poisson($\gamma/j$) new dishes, where $n_i$ is the number of previous customers who have tried $i$th dish.  
\end{itemize}
The aforementioned procedure produces a customer-dish infinite binary matrix $A$ which is known as the feature matrix.

\section{Problem Formulation}
\label{sec:PF}
We consider time-varying multi-object tracking problem where each measurement may or may not be generated from the unknown number of objects. Additionally, not only is the total number of objects unknown but since it is unknown which objects generate each measurement, the number of objects that generate a specific measurement is also assumed to be unknown.

Let us consider a multiple object tracking model with time-varying object cardinality $N_k$ ($N_k$ is unknown) and time-varying observation cardinality $M_k$, at time step $k$. Let $\mathbb{X}_k = \{\bx_{1,k},\dots, \bx_{N_k, k}\}$ and $\mathbb{Y}_k = \{\by_{1,k},\dots, \by_{M_k,k}\}$ be the time varying collection of states and measurements  at time step $k$, respectively. We assume that each measurement may or may not be generated from more than one objects, however, the measurement-to-object association is not known in advance. Transitioning from time $(k-1)$ to $k$, objects may enter (new objects may appear), leave with probability $(1- \text{P}_{\cdot, k \mid k-1})$, or remain in the field of view with probability $\text{P}_{\cdot, k \mid k-1}$ and then transition according to the (probability) transition kernel $p_{{\underline{\theta}}}(\bx_{\cdot,k} | \bx_{\cdot,k-1})$. The objective is to 
\begin{enumerate}
\item provide measurement-to-object association, and 
\item jointly estimate each object's trajectory and object cardinality.
\end{enumerate}
\section{Tracking with spawning framework}
\label{sec:proposed_method}
 We utilize Bayesian nonparametric modeling to place a Dependent Dirichlet process mixture model (DDPMM) introduced in \cite{moraffah2018} on the object states and stated in Section \ref{sec:background}. It is worth noting that our problem setup gives rise to a Bayesian nonparametric setting, where an infinite number of objects may generate each measurement. In particular,  we employ an Indian Buffet Process (IBP) as a prior over the allocation matrix $A$. To do so, we define the infinite binary allocation matrix $A = [a_{i,j}]$ as follows 

\be
a_{i,j} = 
\left\{
	\begin{array}{ll}
		1  & \mbox{if } \text{$i$th measurement is originated} \\
		&\mbox{} \text{ from the $j$th object} \\
		0 & \mbox{ } \text{otherwise}.
	\end{array}
\right.
\ee
\normalsize
Note that we assume matrix $A$ is the infinite latent matrix that determines from which objects each measurement is originated as the number of objects generating each measurement is unknown. It is also worth mentioning that as all Bayesian nonparametric models suggest, only a finite number of elements in matrix $A$ is non-zero \cite{JMLR:v12:griffiths11a}. We use an Indian buffet process (IBP) as a prior over such infinite binary matrices. We then infer the object identities and, hence, successfully track the multiple objects using its corresponding measurements in a robust manner.

\noindent Our proposed model consists of two main parts: 
\begin{enumerate}
\item[(A)] One is a dependent Dirichlet Process (DDP) component which creates a multi-state prior by learning multiple object clusters or labels over related information; 
\item[(B)] an Indian buffet process (IBP) which accounts for the measurement-to-object association.
\end{enumerate}
Integrating (A) with (B), we define a framework that can track multiple time-varying objects using their associated measurements. 

\subsection{Dependent Dirichlet Process Prior on State Transition  }

In this section, we exploit the properties of the Dependent Dirichlet process (DDP) introduced in \cite{moraffah2018, maceachern1999dependent}. Let $G$ be a probability random measure drawn from a DDP($\alpha$). Note that it is shown that DDP is discrete with probability one. Therefore, using DDP as a prior over states is restrictive. We thus employ the dependent Dirichlet mixture model. In particular, let $\mathbb{X}^{\ell-1}_{k}$ and $\Theta^*_{k|k-1}$ be the configuration at time $k$ up to the $\ell$th object and the set of unique parameters survived and evolved to time $k$, respectively, then the generative model is defined as follows: 
\be
\label{eq: mainmodel1}
&G\sim \text{DDP}(\alpha)\notag\\
&\Theta_k | G \sim G\\
&{\bf{x_{\ell,k}}} | {\mathbb{X}}^{\ell-1}_{k}, {\mathbb{X}}^*_{k|k-1}, \Theta^*_{k|k-1}, \Theta_k\sim F\notag
\ee 
where $F$ is a distribution whose probability density function depends on the physical model. 

\subsection{Measurement-to-Object Association }
Measurements can be originated from multiple different objects. However, objects as well as  the number of objects from which a measurement is generated are not known. In this section, we propose a Bayesian nonparametric model to address the association problem. to this end, let $A_k$ be the association matrix where each entry of the matrix at time $k$, where $a^k_{i,j} = 1$ if the $j$th measurement is generated from the $i$th object and $a^k_{i,j} = 0$ otherwise. Note that matrix $A$ is a latent matrix. We define an Indian buffet process as a prior over $A$. It is worth noting that this prior similar to Chinese restaurant process imposes a \textit{"rich gets richer"} effect and thus the matrix is sparse even though it has infinite dimension. To generate the measurements, we define $W_k = \mathbb{X}_k \odot A_k$ where $\odot$ indicates the Hadamard product (element-wise product). The generative model is defined as follows
\be
\label{eq: mainmodel2}
&A_k\sim \text{IBP}(\gamma)\notag\\
&\phi_{j,k}| {G_m}\sim G\\
&\by_{j,k}| \phi_{j,k}, \mathbb{X}_k, A\sim R(\cdot|\phi_{j,k}, \{\bx_{\ell,k}\}_{\{\ell:a_{j,\ell}\neq 0\}}, W_k)\notag
\ee
where $G$, $F$, and $R$ are some distribution following the physical model. One should determine the hyperparameters $\alpha$ and $\gamma$, in order to accurately estimate the trajectory of each object. We take a Bayesian approach and use Gamma  distributions as priors over the hyperparameters. 

\section{Inference}
In this section, we aim to estimate the latent variables as well as the object trajectory, i.e., the state variables at each time $k$, conditioned on the observations up to time $k$. It is worth mentioning we are only interested in the state estimation and the rest of parameters and latent variables are nuisance. Nonetheless, for the sake of completeness, we provide the full posterior distribution. Marginalizing out is simple due to IBP representation of the model. The posterior distribution of interest is $\mathbb{P}(\{{\bf{x_{\ell,k}}}\}_{k, \ell}, \Theta_k, \Phi_k, W_k |\{\mathbb{Y}_k\}_k, \alpha, \gamma)$, where $\Theta_k, \Phi_k$ are the set of parameters for the model at time $k$. In particular, we are interested in 
\begin{multline}
\label{eq:posterior}
\mathbb{P}(\{{\bf{x_{\ell,k}}}\}_{k, \ell}, \Theta_k, \Phi_k, W_k |\{\mathbb{Y}_k\}_k, \alpha, \gamma) \propto \\
\mathbb{P}({\bf{x_{\ell,k}}}, \Theta_k, \Phi_k, A_k , \{\mathbb{Y}_k\}_k| \alpha, \gamma)\\
 = \mathbb{P}(A_k|\gamma)\prod_{k=1}^{K}\bigg[\prod_{\ell = 1}^{D_k}\mathbb{P}(\theta_{\ell,k}|\theta_{\ell, k-1}, \alpha)\\
 \prod_{j=1}^{M_k}\prod_{\ell=1}^{N_k}\mathbb{P}({\bf{x_{\ell,k}}}|{\bf{x_{\ell,k-1}}}, \theta_{\ell,k})\mathbb{P}(\phi_{j,k}) \\
 \big[\mathbb{P}(y_{j,k}|\phi_{j,k}, \{{\bf{x_{\ell,k}}}\})\big]^{\mathbbm{1}(\ell:a^k_{j,\ell}\neq 0)}\bigg]
\end{multline}
where $\mathbbm{1}(\cdot)$ is an indicator function; i.e., the indicator function is $1$ if the argument is true and otherwise it is zero. Note that all terms in the posterior equation (\ref{eq:posterior}) are defined through equations (\ref{eq: mainmodel1}) and (\ref{eq: mainmodel2}). Computing the exact posterior mean is impossible as there is no closed form solution for the integral. Thus, we take advantage of Markov chain Monte Carlo methods. In this paper, we employ the collapsed Gibbs sampling technique to sample from the posterior distribution.

\section{Experiment}
\label{sec:ex}
\begin{figure*}[th]
	\centering
	\hspace*{-.4in}
	\subfloat[ ]{\label{figa}
		\resizebox{2.6in}{!}{\includegraphics{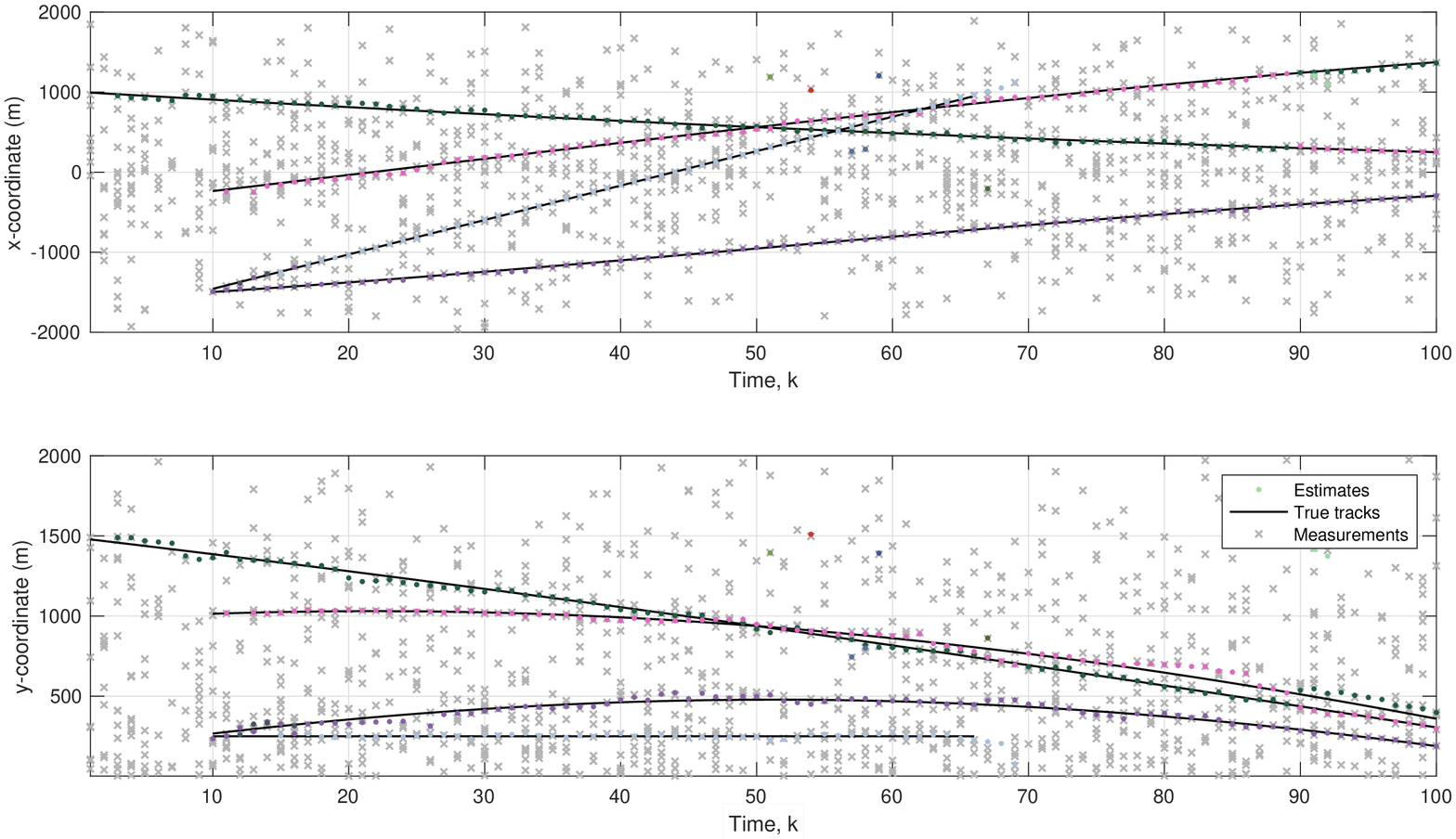}}} 
	\hspace*{-.31in}
	\subfloat[ ]{\label{figb}
		\resizebox{2.6in}{!}{\includegraphics{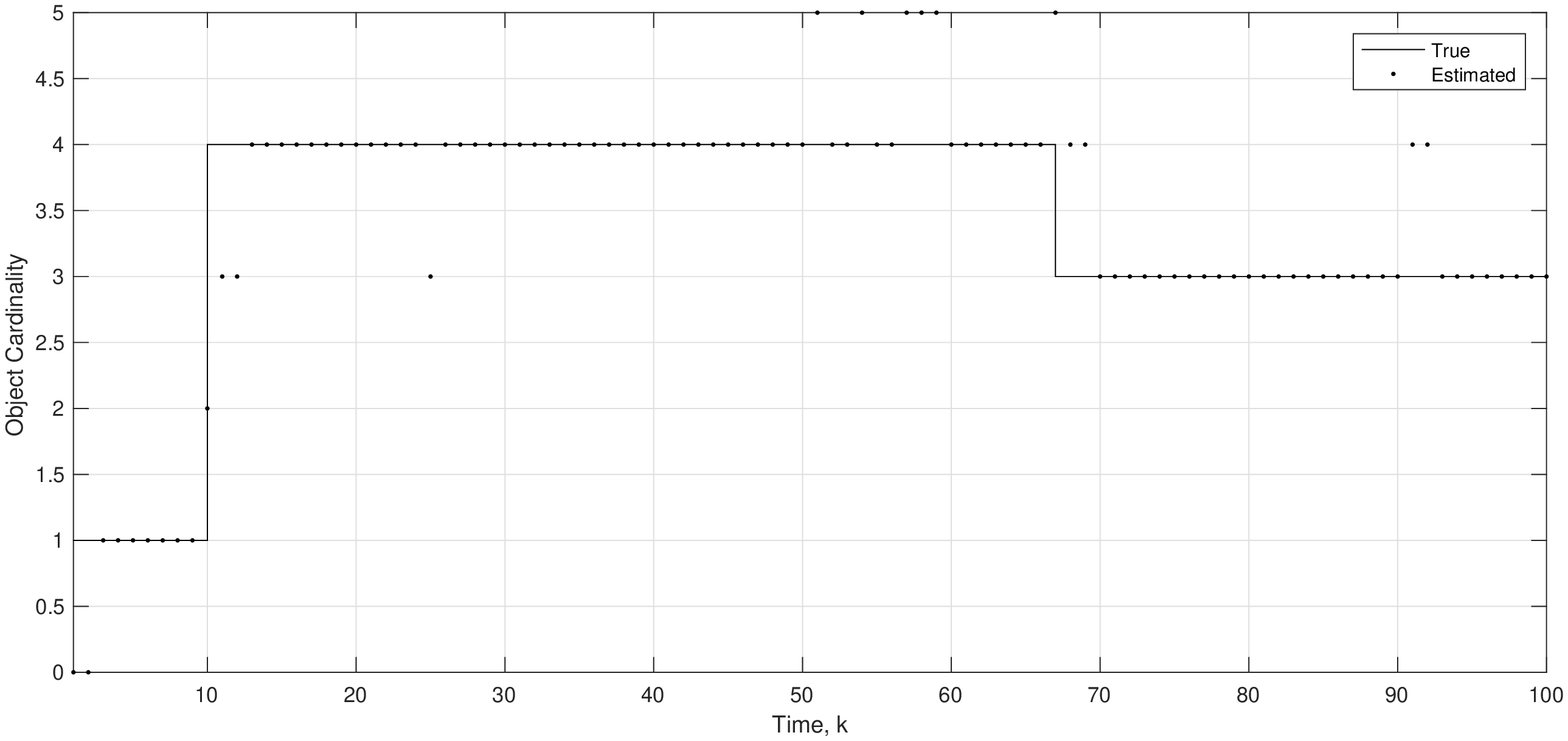}}} 
	\hspace*{-.3in}
	\subfloat[ ]{ \label{figc}
		\resizebox{2.6in}{!}{\includegraphics{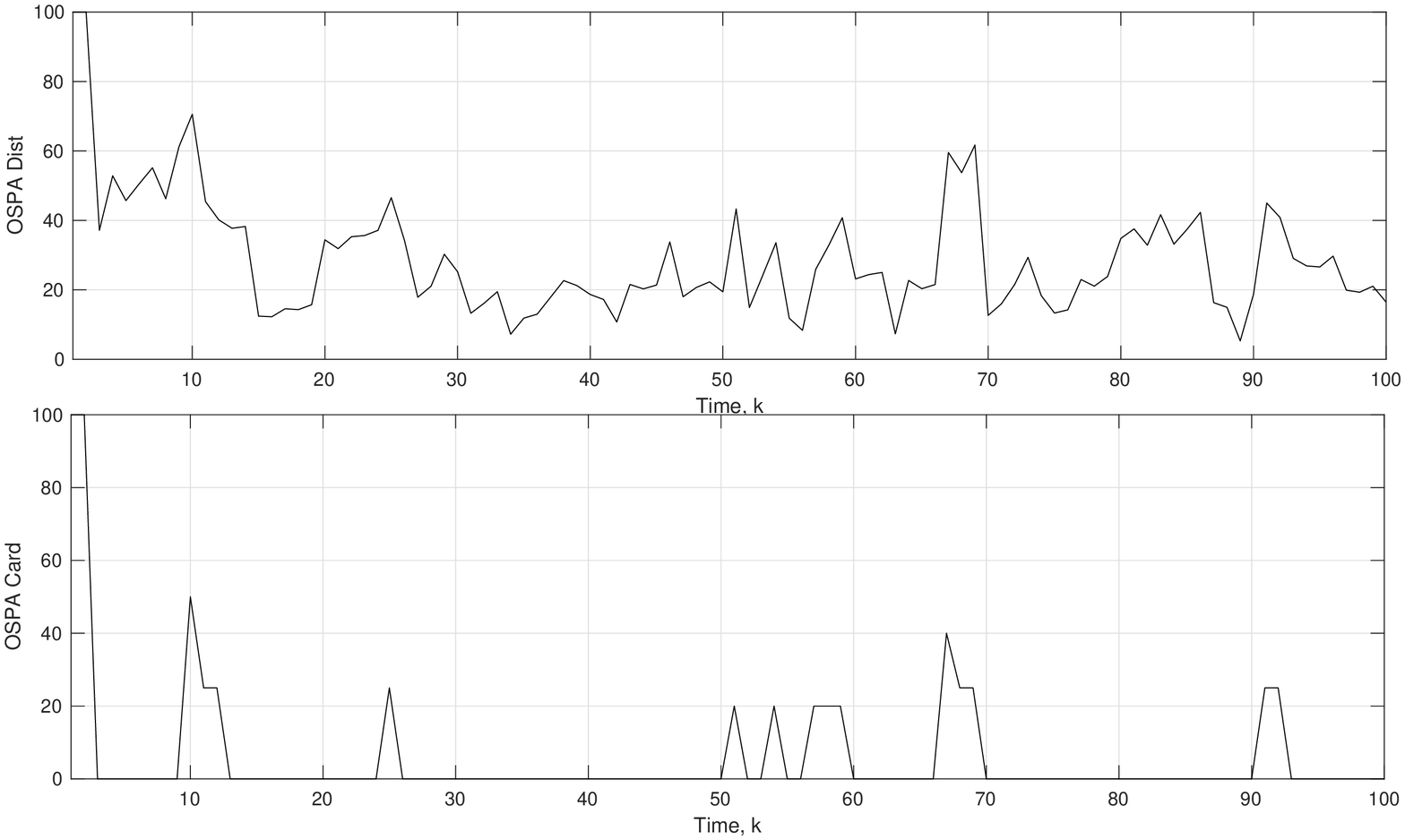}}}
	\vspace*{-3mm}
	\caption{(a) True and estimated x-coordinate and y-coordinate using the proposed method for 4 objects; (b) Cardinality estimation through the proposed method; (c) OSPA metric for the location and cardinality for order $p=1$ and cut-off $c=100$. The results are by 1000 Monte Carlo runs.}
	\label{fig}
\end{figure*}
\subsection{Experiment I: Performance of Proposed Method}
\label{subsec:ex1}
The performance of the proposed framework is demonstrated through experiments.  To this end, we use the same model as \cite{Bryant18}.  In this model, number of objects varies due to birth and death of the objects. To this end, the state model is an autonomous linear Gaussian with state vector $\bx = [x, y, v_x, v_y]^T$ where the first two coordinates denote the $x$ and $y$- coordinate of the object position and the last two coordinates are the object velocity following:
\be
\bx_k = A \bx_{k-1} + \zeta_k, \hspace{1cm} \zeta_k\sim\mathcal{N}(0, \Sigma_s)
\ee
where 
\[
A = \begin{bmatrix}
I_2 & \Delta I_2\\
0_2 &I_2
\end{bmatrix} \hspace{0.75cm}\text{and} \hspace{0.75cm}
\Sigma_s = \sigma_s^2\begin{bmatrix}
\frac{\Delta^4 }{4} I_2& \frac{\Delta^3 }{2} I_2\\
\frac{\Delta^3 }{2} I_2 & \Delta^2  I_2
\end{bmatrix}.
\]
We set $\Delta = 1$s, $\sigma_s = 1$m$\text{s}^{-2}$, $I_2$, and $0_2$ are $2\times 2$ identity and zero matrices, respectively. The observations are also assumed to follow a linear Gaussian equation:
\be
\by_k = B \by_{k-1} + \eta_k, \hspace{1cm} \eta_k\sim\mathcal{N}(0, \Sigma_o),
\ee
where 
\[
B = \begin{bmatrix}
I_2 & 0
\end{bmatrix} \hspace{0.75cm}\text{and} \hspace{0.75cm}
\Sigma_s = \sigma_o^2 I_2, \hspace{0.2cm} \sigma_o^2 = 10 m. \]
For the sake of this simulation, we assume all the densities in our framework are too Gaussian. We also placed a Normal-inverse-Wishart on the parameters of Gaussian distribution as $\text{NIW}(\mu_0, 0, \nu, I)$ for $\mu_0 = 0.001$ and $\nu = 50$ and Gamma distribution priors  on our hyperparameters-- concentration parameter $\alpha$-- as Gamma$(\alpha;  1, 0.1)$. In  Figure (\ref{figa}), we display the $x$ and $y$- coordinate range as well as the estimation of them through the proposed method. 

The OSPA metric of order $p=1$ with cut-off $c=100$ for the proposed method is depicted in Figure (\ref{figb}). We also successfully estimate the time-varying cardinality of objects which is shown in Figure (\ref{figc}). To estimate the trajectory we the associations we employ collapsed Gibbs sampling techniques and run 1000 Monte Carlo runs.

\subsection{Experiment II: Comparison to GLMB with Spawning}

\begin{figure}[h]
    \centering
    \includegraphics[width=0.5\textwidth]{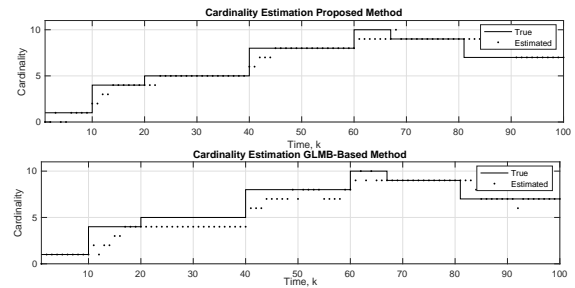}
    \caption{True and estimated object cardinality at each time step using DDP-bases and GLMB-based algorithms for 1000 Monte Carlo runs.}
    \label{fig:card_comp}
\end{figure}

In the previous section, we demonstrated the performance of our proposed method. In this section, we provide a comparison between our proposed work and the GLMB-based spawning technique. In order to provide a fair comparison, we used the same setup introduced in \cite{Bryant18}. This experiment is performed to demonstrate the advantage of our proposed framework over the GLMB-based technique. In Figure \ref{fig:traj_comp}, we show the $x$ and $y$ coordinates of the range as well as an estimation of them through the proposed method. Figure \ref{fig:card_comp}, however, compare the cardinality estimation through our proposed method versus the cardinality estimation using GLMB-based algorithms. In Figure \ref{fig:traj_comp}, the OSPA comparison between the two methods is provided. As depicted, under the same conditions, the proposed method outperforms the existing GLMB-based spawning technique. The OSPA metric for comparison is provided for order $p=1$ with cut-off $c=100$. The comparison between the Bayesian nonparametric modeling and the GLMB-based technique verifies the advantage of the former. 

\begin{figure}[h]
    \centering
    \includegraphics[width=0.5\textwidth]{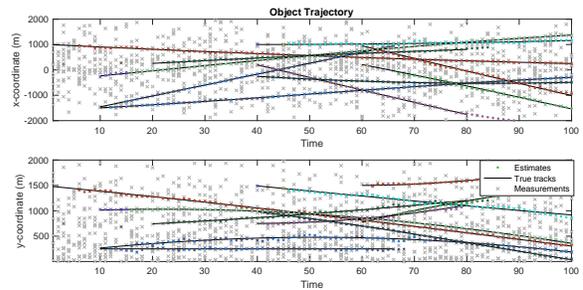}
    \caption{True and estimated object trajectory for 10 objects using our proposed method based on DDP.}
    \label{fig:traj_comp}
\end{figure}

\begin{figure}[h]
    \centering
    \includegraphics[width=0.5\textwidth]{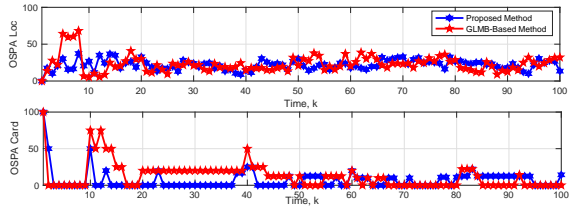}
    \caption{OSPA metric for both cardinality and the location for p = 1, c = 100.}
    \label{fig:traj_comp}
\end{figure}

\section{Conclusion} 
In this paper, we introduced a class of robust Bayesian nonparametric models to address multi-object tracking with spawning. Our proposed model has two folds. The first fold exploits the Dependent Dirichlet process to account for the birth and death of objects and the second fold utilizes the Indian Buffet Process as a prior over the association matrix for which one can associate each measurement to the corresponding objects. Integrating with a Markov chain Monte Carlo sampling technique -- in this paper, we used collapsed Gibbs sampling-- to sample from the posterior distribution. This framework accurately and robustly provides objects' trajectory estimation, object cardinality as well as measurement-to-object association. We demonstrated the performance of our proposed framework through simulations and showed that this framework, under the same conditions,  may outperform the RFS-based techniques.

\bibliographystyle{IEEEtran}
\bibliography{ref_BM.bib}

\end{document}